\documentclass[letterpaper]{article} %
\usepackage{aaai2026}  %
\usepackage{times}  %
\usepackage{helvet}  %
\usepackage{courier}  %
\usepackage[hyphens]{url}  %
\usepackage{graphicx} %
\usepackage{natbib}  %
\usepackage{caption} %
\usepackage{algorithm}
\usepackage{algorithmic}

\usepackage{url}            %
\usepackage{booktabs}       %
\usepackage{amsfonts}       %
\usepackage{amsfonts}       %
\usepackage{amsmath}        %
\usepackage{nicefrac}       %
\usepackage{microtype}      %
\usepackage{xcolor}         %

\usepackage{colortbl}
\usepackage{graphicx}
\usepackage{marvosym}
\usepackage{caption}
\usepackage{subcaption}
\usepackage{graphicx}

\usepackage{bbding}
\usepackage{amssymb}
\usepackage{wasysym} %

\usepackage{newfloat}
\usepackage{listings}
\DeclareCaptionStyle{ruled}{labelfont=normalfont,labelsep=colon,strut=off} %
\floatstyle{ruled}
\newfloat{listing}{tb}{lst}{}
\floatname{listing}{Listing}
\nocopyright

\title{MagicVL-2B: Empowering Vision-Language Models on Mobile Devices with Lightweight Visual Encoders via Curriculum Learning}

\author{
    Yi Liu\equalcontrib \thanks{Project Leader, \textsuperscript{\Letter}Corresponding Author.} ,
    Xiao Xu\equalcontrib,
    Zeyu Xu\equalcontrib,
    Meng Zhang\equalcontrib,
    Yibo Li\equalcontrib, 
    Haoyu Chen\equalcontrib, \\
    Junkang Zhang, Qiang Wang, Jifa Sun, Siling Lin,\\
    Shengxun Cheng, Lingshu Zhang, Kang Wang \textsuperscript{\Letter} }

\affiliations{
    Honor Device Co., Ltd~~~\\
    wangkang12@honor.com
}

\begin{document}

\maketitle

\begin{abstract}
Vision-Language Models (VLMs) have achieved remarkable breakthroughs in recent years, enabling a diverse array of applications in everyday life. However, the substantial computational and storage demands of VLMs pose significant challenges for their efficient deployment on mobile devices, which represent the most ubiquitous and accessible computing platforms today. In this work, we introduce \textbf{MagicVL-2B}, a novel VLM meticulously optimized for flagship smartphones. MagicVL-2B leverages a lightweight visual encoder with fewer than 100M parameters and features a redesigned dynamic resolution scheme that adaptively generates image tokens without excessive modification of image dimensions. To further enhance the performance of this compact encoder within VLMs, we propose a multimodal curriculum learning strategy that incrementally increases task difficulty and data information density throughout training. This approach substantially improves the model’s performance across a variety of sub-tasks. Extensive evaluations on standard VLM benchmarks demonstrate that MagicVL-2B matches the accuracy of current state-of-the-art models while reducing on-device power consumption by \textbf{41.1\%}. These results establish MagicVL-2B as a practical and robust solution for real-world mobile vision-language applications, enabling advanced multimodal intelligence to run directly on smartphones.
\end{abstract}
\section{Introduction}\label{introduction}

\begin{figure*}[t]
\centering
\includegraphics[width=\textwidth]{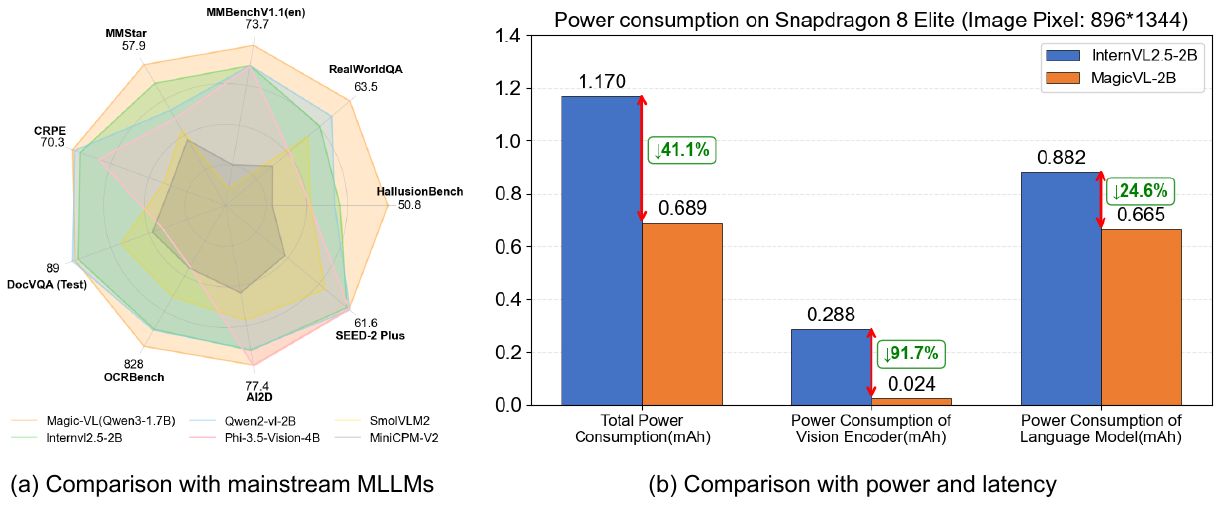}
\caption{
\textbf{Comparison between MagicVL-2B and mainstream VLMs.}
(a) MagicVL-2B demonstrates competitive performance across a wide range of multimodal benchmarks, matching or even surpassing other small-scale models. 
(b) MagicVL-2B achieves substantially lower inference power consumption and latency compared to InternVL2.5-2B, underscoring its efficiency and practicality for real-world deployment.
}
\label{fig:main_results}
\end{figure*}

In recent years, Vision-Language Models (VLMs)~\cite{achiam2023gpt,chen2024internvl,chen2024far,bai2023qwen,wang2024qwen2,mckinzie2024mm1,zhang2024mm15methodsanalysis,tong2024cambrian} have achieved remarkable breakthroughs, enabling a wide range of real-world applications. These advances have empowered richer human-computer interactions and deeper contextual understanding, resulting in more intuitive and intelligent user experiences. However, the substantial computational and memory demands of VLMs pose a significant barrier to their seamless deployment on mobile devices—the most ubiquitous and user-friendly computing platforms today~\cite{ding2024enhancing,qu2024mobile,hua2024interactivespeculativeplanningenhance,yao2024minicpm,xue2024powerinfer}.

Among all computing platforms, smartphones are especially well-positioned to benefit from VLMs, as they support real-time on-device inference, enabling instant interactions and enhanced privacy~\cite{ding2024enhancing,qu2024mobile}. Deploying VLMs on mobile devices also greatly improves model accessibility, allowing users to conveniently access advanced multimodal models in daily scenarios such as augmented reality, real-time translation, and smart assistants~\cite{hua2024interactivespeculativeplanningenhance,chu2023mobilevlm}. 

Despite these advantages, deploying VLMs efficiently on smartphones remains challenging. First, limited memory capacity restricts the deployment of large-scale models, affecting their representational power and accuracy. Second, the constrained computational capability of mobile processors limits inference speed and energy efficiency. Third, mainstream VLMs typically adopt large Vision Transformer (ViT) encoders, which, due to hardware constraints, result in higher power consumption for visual encoding on-device compared to GPUs in the cloud. Few works leverage lightweight vision encoders to reduce on-device power consumption, likely because such encoders are more difficult to align with Large Language Model (LLM) capabilities, leading to suboptimal performance.

To address these challenges, we present \textbf{MagicVL-2B}, an innovative VLM specifically optimized for flagship smartphones. In terms of algorithmic design, MagicVL-2B employs a lightweight visual encoder tailored for efficient on-device inference, with fewer than 100M ViT parameters, as illustrated in Figure~\ref{fig:framework}(a). This significantly reduces the power consumption of visual encoding on mobile devices. To further unleash the potential of lightweight encoders, we curate a large-scale multimodal dataset and introduce a curriculum learning strategy. By progressively increasing the information density and task difficulty during training, we substantially enhance the capabilities of VLMs with lightweight visual encoders, while maintaining fast and low-power inference.
As shown in Figure~\ref{fig:main_results}, MagicVL-2B achieves both state-of-the-art performance and superior efficiency compared to existing lightweight VLMs. 
Specifically, MagicVL-2B consistently outperforms or matches other small-scale models on a wide range of challenging multimodal benchmarks, while significantly reducing \textbf{41.1\%} inference total power consumption. 
This remarkable combination of accuracy and efficiency highlights MagicVL-2B as a highly practical and scalable solution for real-world multimodal applications, where both resource constraints and model capability are critical requirements.

\textbf{Our main contributions are summarized as follows:}
\begin{itemize}
\item \textbf{Efficient and Lightweight Visual Encoder}: We adopt Siglip2-Base-384/16~\cite{tschannen2025siglip} as an efficient and lightweight visual encoder with fewer than 100M parameters. This encoder is capable of processing images at arbitrary resolutions while producing a compact set of tokens, without modifying the original image size.
\item \textbf{Curriculum Learning Strategy}: We introduce a curriculum learning strategy that systematically structures the training process by staging both information density and task difficulty. By progressively increasing the complexity of training samples and tasks, our approach enables the model to acquire foundational capabilities before addressing more challenging scenarios. This staged progression facilitates more stable convergence and yields significant improvements in overall model performance.

\item \textbf{Superior Performance and Efficiency}: MagicVL-2B achieves state-of-the-art results among models with similar parameter scales, demonstrating superior accuracy across a range of vision-language benchmarks. Furthermore, our model reduces power consumption on mobile devices by 41.1\%, making it highly suitable for real-world on-device applications where efficiency is critical.
\end{itemize}
\section{Related Works}\label{relate}
\subsection{Efficient Image Encoding}

CLIP-pretrained~\cite{CLIP} vision transformers~\cite{ViT} remain the mainstream image encoders for VLMs, with models such as SigLIP~\cite{zhai2023siglip}, EVA-CLIP~\cite{EVA-CLIP}, InternViT~\cite{chen2023internvl}, and DFN-CLIP~\cite{DFN} widely used. Recent works~\cite{karamcheti2024prismatic, tong2024cambrian1, shi2024eagle} improve performance by ensembling visual encoders with diverse objectives, while methods like LLaVA-PruMerge~\cite{shang2024LLaVA-PruMerge} and Matryoshka-based token sampling~\cite{hu2024matryoshka, cai2024matryoshka} dynamically prune visual tokens to improve encoding efficiency. Additional strategies~\cite{instructblip, cha2023honeybee, chu2023mobilevlm, chu2024mobilevlm} leverage perceiver-style resamplers or pooling operations to reduce token numbers. Hierarchical architectures such as ConvNeXT~\cite{liu2022convnet} and FastViT~\cite{vasu2023fastvit} further decrease token counts via downsampling the input tensor at each computational stage.

\subsection{Vision Language Models}
LLMs~\cite{brown2020language, touvron2023llama1, touvron2023llama2, anil2023palm} have proven highly effective in tackling a wide spectrum of challenging tasks~\cite{wei2022emergent, trinh2024solving}. Vision-language models (VLMs)~\cite{Achiam2023GPT4TR, liu2023llava, chen2024internvl, zhang2023llamaadapter} extend LLMs to multimodal inputs via mechanisms such as linear projectors~\cite{liu2023llava, chen2024far, wang2024qwen2}, Q-Former modules~\cite{li2023blip}, and perceiver resamplers~\cite{alayrac2022flamingo, bai2023qwen,yao2024minicpm}. To better process high-resolution images, dynamic resolution techniques~\cite{chen2024far,liu2023improvedllava,liu2024llavanext} have been introduced, enabling finer-grained visual understanding at varying resolutions~\cite{huang2024mini}. However, these dynamic resolution methods introduce specific challenges for mobile deployment: the proliferation of image patches can substantially slow down the visual encoder, while the resulting longer sequences of image tokens lead to increased inference latency for the language model~\cite{lin2023vila}.


\subsection{On-Mobile-Device Large Language Models}
With the expansion of application scenarios for large language models, there is growing interest in small-scale large language models (SLMs) as users prioritize efficiency and cost-effectiveness~\cite{ashkboos2024computational}. Recently, a range of SLMs have been developed to address these needs, covering both language-only~\cite{hu2024minicpm,abdin2024phi,mehta2024openelm} and multimodal~\cite{yao2024minicpm,wang2024qwen2,chen2024far,luo2024mono,li2024llava} models. Thanks to their reduced parameter counts (typically 2-3B), these models are now feasible for deployment on personal devices such as PCs and smartphones. Beyond the creation of more compact yet powerful LLMs and VLMs, recent system-level research has proposed various approaches for efficiently deploying SLMs on end-user hardware, including personal computers~\cite{wei2024t} and mobile phones~\cite{yao2024minicpm,li2024transformer}. 
Our proposed MagicVL-2B adopts a smaller visual encoder, which enables the model to achieve significantly lower power consumption on mobile devices, while still delivering strong performance across a wide range of benchmarks.

\begin{figure*}[t]
\centering
\includegraphics[width=\linewidth]{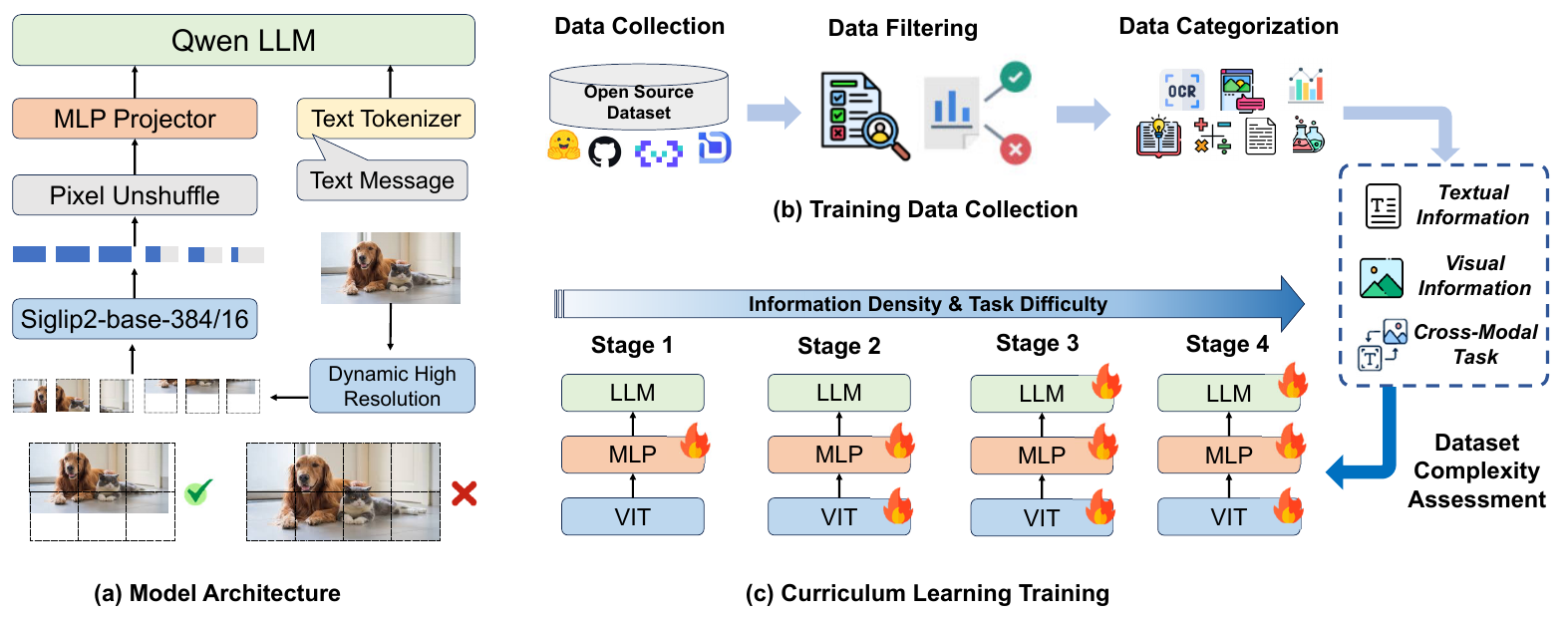}
\caption{
Overview of our MagicVL-2B. 
(a) \textbf{Model Architecture}: The model architecture integrates large language model using a visual encoder  with dynamic high resolution and MLP projector.
(b) \textbf{Training Data Collection}: Open-source datasets are filtered and categorized into sub-tasks based on data quality and task types.
(c) \textbf{Curriculum Learning Training}: Dataset complexity is assessed along three dimensions. The model is trained in multiple stages, with each stage introducing tasks of increasing difficulty and more complex information. 
}
\label{fig:framework}
\end{figure*}

\section{MagicVL-2B}\label{method}

In this section, we provide a comprehensive overview of MagicVL-2B, focusing on its model architecture and the design of a lightweight visual encoder. We also describe how a curriculum learning strategy is employed to progressively train the model.

\subsection{Model Architecture }

\subsubsection{Overall Architecture}
Our architecture is an enhanced variant built upon the InternVL2.5 framework~\cite{chen2024far}. The overall pipeline is depicted in Figure~\ref{fig:framework} and comprises the following key components. \textbf{Visual Encoder}: To handle multimodal (image and language) inputs, we employ the SigLIP2-base~\cite{tschannen2025siglip} Vision Transformer (ViT) with an input resolution of 384$\times$384, as adopted in prior works~\cite{lin2023vila}. This encoder contains 93M parameters. \textbf{MLP Projector}: A two-layer multilayer perceptron (MLP) is utilized to project image tokens into the token space of the large language model (LLM). \textbf{LLM}: We leverage Qwen2.5-1.5B~\cite{qwen2.5} or Qwen3-1.7B~\cite{qwen3} as the backbone language model to construct MagicVL-2B. To further enhance the model's capability in comprehending inputs at varying resolutions, we introduce a Dynamic High Resolution module. Inspired by the limitations of excessively enlarged images observed in InternVL2.5~\cite{chen2024far} and LLaVA-NeXT~\cite{liu2024llavanext}, we propose a novel solution that significantly improves both training and inference efficiency.

\subsubsection{Lightweight Visual Encoder}
The visual encoder in our model is responsible for processing image-modality inputs. We systematically evaluated three candidate architectures: ViT~\cite{ViT}, SigLIP~\cite{zhai2023siglip}, and SigLIP2~\cite{tschannen2025siglip}, and ultimately selected SigLIP2 due to its superior performance. To achieve an optimal trade-off between computational efficiency and the capability to extract fine-grained visual features for on-device VLMs, we adopt the following design choices: First, regarding model size, we employ the SigLIP2-base variant, which comprises approximately 93 million parameters, to maintain computational efficiency. Second, for image resolution, we use a relatively high input resolution (384$\times$384) to improve the representation of global visual information. Third, we set the patch size to 16, enabling the encoder to better capture fine-grained and complex visual details, which is particularly beneficial for on-device scenarios.

\subsubsection{Dynamic High Resolution}
Our visual encoder is pre-trained on a fixed input resolution of 384, which significantly constrains its adaptability to various images, especially higher resolutions. Dynamic resolution has emerged as an effective approach to address this limitation, as demonstrated by models such as InternVL~\cite{zhu2025internvl3,chen2024expanding}. However, existing dynamic resolution techniques often suffer from severe distortion artifacts, typically requiring resizing the original image's height and width to integer multiples of the pre-training resolution. As a result, when the aspect ratio of the input image differs from that of the pre-training resolution, the image content is easily distorted, leading to degraded semantic representation and the introduction of redundant tokens, which further reduces inference efficiency. This issue is particularly evident on mobile devices, where atypical aspect ratios (e.g., long screenshots) are prevalent.

To address this problem, we propose a token-level resizing strategy: instead of resizing the image to integer multiples of the pre-training resolution, we resize each dimension to the nearest integer multiple of the pixel size corresponding to a single visual token. This approach minimizes image distortion under the VLM token paradigm, ensuring that image resolution and content are almost perfectly preserved.
Given an input image
$
V \in \mathbb{R}^{H \times W \times C}
$
where $H$, $W$, and $C$ denote the original image height, width, and number of channels, respectively, the resized image is
$
V' \in \mathbb{R}^{H' \times W' \times C}
$
where $H'$ and $W'$ are computed as:
\[
\begin{array}{ccc}
H' = \left\lfloor \frac{H}{N_{\text{token}}} + 0.5 \right\rfloor \times N_{\text{token}} \\
W' = \left\lfloor \frac{W}{N_{\text{token}}} + 0.5 \right\rfloor \times N_{\text{token}} \\
N_{\text{token}} = N_{\text{patchsize}} \times R_{\text{psf}}
\end{array}
\]
Here, $\lfloor \cdot \rfloor$ denotes the floor function, $N_{\text{patchsize}}$ is the patch size (set to 16).
We utilize pixel unshuffle for token compression, with a compression ratio of $R_{\text{psf}} = 2$.
To accommodate images of varying sizes with an encoder that operates at a fixed input resolution, we standardize all inputs to match the encoder’s required resolution. To prevent image distortion during this unification process, we adopt a {padding} strategy: for any image boundary $v_i$ that does not meet the required dimension, we pad the image with zeros until it reaches $384 \times 384$. The influence of these padded regions is eliminated by applying an {attention mask}.
All tokens generated from the padded regions are discarded, and only tokens corresponding to the original image content are retained for subsequent LLM computation. This approach maximizes the preservation of the original image information while minimizing the introduction of redundant information during the size unification process.

\subsection{Training Data Collection}


\subsubsection{Data Collection}
For pre-training, we curated a large-scale collection of open-source image-text datasets, comprising approximately \textit{150 million image-text pairs}. We prioritized datasets that offer both high data quality and diverse visual content. For datasets that feature substantial visual diversity but only moderate quality, we applied rigorous filtering and data cleaning procedures to improve their overall reliability. Due to the limited availability of open-source Chinese image-text datasets, we leveraged large-scale LLMs to translate a subset of English datasets into Chinese, thereby enhancing the model's bilingual capabilities. A comprehensive description of our datasets collection can be found in the supplementary materials.

\subsubsection{Data Filtering}
We employ a multi-stage data filtering pipeline to ensure high-quality image-text pairs for pre-training. First, a \textit{heuristic rule-based filtering system} is applied to remove samples containing excessive abnormal characters or synthetic data with anomalous keywords. To mitigate the issue of repetitive content observed in InternVL2.5~\cite{chen2024expanding}, we further design a \textit{rule-based duplication detection system} that eliminates entries with large repeated segments or frequent occurrences of short phrases. Finally, we introduce an \textit{LLM prompt-based filtering system}, which leverages large language models to evaluate the \textit{logical coherence} of each entry and to detect potential \textit{hallucinations} in the descriptions. Representative examples of excluded data can be found in the supplementary materials.


\subsubsection{Data Categorization}

We systematically organize the collected open-source datasets into the following task categories: \emph{reasoning}, \emph{GUI}, \emph{OCR}, \emph{text-only}, \emph{chart}, \emph{caption}, \emph{visual question answering}, and \emph{grounding}. To ensure accurate and efficient categorization, we first inspect each dataset to determine whether explicit task-type labels are provided. If such labels exist, we directly categorize the dataset accordingly. For datasets lacking explicit task-type labels, we conduct manual verification by randomly sampling a subset of data points and performing human inspection to assess whether the dataset corresponds to a single task category. If multiple task types are present within a dataset, we further employ a large-scale vision-language model (VLM) to automatically classify individual samples, thereby splitting heterogeneous datasets into several task-specific subsets.

\subsection{Curriculum Learning Training}

\subsubsection{Dataset Complexity Assessment}

Given a dataset $\mathcal{D}$, where $v$ denotes the input image, $p$ the input prompt, $x$ the corresponding response, and $n$ the total number of samples, we represent the dataset as
$$
\mathcal{D} = \{(v_i, p_i, x_i) \mid i = 1, \ldots, n\}.
$$
As depicted in Figure~\ref{fig:framework}(c), we introduce a rigorous and multifaceted evaluation protocol for characterizing dataset complexity. 
Our framework systematically dissects the dataset along three different dimensions, yielding a holistic complexity score $S$ that encapsulates the textual information, visual information, and cross-modal task complexity.


\subsubsection{Textual Information Complexity} This dimension evaluates the complexity of a dataset based on the diversity and linguistic complexity of the textual content.
    \begin{enumerate}

    \item  \emph{Token length:} We define the normalized average token length of response $L$ as an indicator of the level of detail and informativeness:
    $$
    L = \frac{1}{n}\sum_{i=1}^n \text{len}(x_i),
    $$
    where $\text{len}(x_i)$ denotes the token length of $x_i$.
    
    \item \emph{Type-token ratio (TTR):} To capture lexical diversity, we calculate the average type-token ratio for the concatenation of prompt and response:
    $$
    T = \frac{1}{n}\sum_{i=1}^n \text{TTR}(p_i+x_i),
    $$
    where a higher TTR indicates greater lexical diversity, reflecting increased linguistic complexity\cite{kettunen2014can}.
    
    \item \emph{Perplexity:} Besides the statistical methods, we also utilize the language model in our model architecture, Qwen2.5-1.5B, to compute the average perplexity of the response $x_i$ conditioned on prompt $p_i$: 
    $$
    P = \frac{1}{n}\sum_{i=1}^n\text{PPL}(x_i \mid p_i).
    $$
    The perlexity produced by the language model not only reflect the intrinsic linguistic complexity of the textual data~\cite{ankner2024perplexed}, but also implicitly capture the degree to which the text content is dependent on or grounded in the image data, providing an additional perspective for measuring dataset complexity.
    \end{enumerate}
    The overall text-based complexity score $S_{text}$ is then given by the arithmetic mean after normalization:
    $$
    S_{\text{text}} = \frac{1}{3} \left( \hat{L} + \hat{T} + \hat{P} \right),
    $$
    where the notation $\hat{\cdot}$ denotes normalization. 
    
\subsubsection{Visual Information Complexity} This dimension assesses the complexity of a dataset by quantifying the richness of visual contents in the images.
    \begin{enumerate}
    \item \emph{Image entropy:} we calculate the average image entropy as a statistical measure of the pixel-level information content. Let $e$ denote the normalization constant for entropy, and we define our average \emph{image entropy} $E$ of the dataset as:
    $$
     E =  \frac{1}{n}\sum_{i=1}^n \text{Entropy}(v_i).
    $$
    
    \item \emph{Text density:} we utilize a state-of-the-art OCR model to compute the average text density within images in the dataset~\cite{cui2025paddleocr30technicalreport}. Specifically, Let $t(v_i)$ denote the number of text tokens recognized by the OCR model in image $v_i$, and $a_i(v_i)$ denote the area (in pixels) of image $v_i$. The average text density per image area is calculated as
    $$
    D_{\text{text}} = \frac{1}{n} \sum_{i=1}^n \frac{t(v_i)}{a_i(v_i)}.
    $$
    
    \item \emph{Object density:} similarly, we also leverage an open-domain object detection model to estimate the average object density of each image~\cite{liu2023grounding}. Let $\text{obj}(v_i)$ denote the number of objects detected in image $v_i$, then the average object density per image area is defined as:
    $$
    D_{\text{obj}} = \frac{1}{n} \sum_{i=1}^n \frac{\text{obj}(v_i)}{a_i(v_i)}.
    $$
    \end{enumerate}
    We normalize each metric and compute the overall image-based complexity score $S_{image}$ as:
    $$
    S_{\text{image}} = \frac{1}{3} \left( \hat{E} + \hat{D}_{\text{text}} + \hat{D}_{\text{obj}} \right).
    $$
    
\subsubsection{Cross-Modal Task Complexity} This dimension assesses dataset complexity in terms of cross-modal information integration and reasoning. Following NVILA~\cite{nvila}, we propose a loss-based framework that uses VLMs of different scales to quantify task complexity. The key idea is that samples demanding more advanced cross-modal reasoning will yield a larger gap in autoregressive loss between smaller and larger models.
    
    \emph{Loss-based comparative evaluation:}  Let $M_s$ and $M_l$ denote a small and a large VLM, respectively. For each data point, we compute the autoregressive loss of VLM models, $M_{s}(x \mid (v, p))$ and $M_{l}(x\mid (v, p))$. We then define the complexity score as the proportion of data points where the loss gap between $M_s$ and $M_l$ exceeds a controlled margin:
    $$
    \mathcal{C}(M_s, M_l) = \frac{1}{n} \sum_{i=1}^n \mathbf{1}_{\{M_s(x_i \mid(v_i, p_i)) > \beta M_l(x_i \mid (v_i, p_i)) > \delta \}},
    $$
    where $\beta$ is a scaling hyperparameter determining the required loss ratio, and $\delta$ is a threshold to filter out trivial cases with low absolute loss. For a more comprehensive and robust evaluation, we calculate the autoregressive loss for three models with different sizes: Qwen2VL-2B, Qwen2VL-7B, and Qwen2VL-72B~\cite{wang2024qwen2}, and define our cross-model task complexity score $S_{\text{task}}$ as:
    $$
    S_{\text{task}} = \frac{1}{2} \left(C(M_{2B}, M_{7B}) + C(M_{7B}, M_{72B}) \right).
    $$
    
With the three complexity scores $S_{\text{text}}, S_{\text{image}}, S_{\text{task}}$ obtained from the aforementioned metrics, we compute the overall dataset complexity score $S$ as a weighted sum:
$$
S = \lambda_1 S_{\text{text}} + \lambda_2 S_{\text{image}} + \lambda_3 S_{\text{task}},
$$
where the weights $\{\lambda_i\}_{i=1}^{3}$ are adaptively selected according to the task category of each dataset, as defined in the previous section. Detailed weight configurations for each task category are provided in supplementary materials.
\begin{table*}[t]
\centering
\begin{tabular}{lcccccccccc}
\toprule
Model & HB & CR & MMB & RQA & MME\_R & MMS & DocV & OB & A2D & SEED \\
\midrule
\multicolumn{11}{c}{\textbf{Model parameters $>$ 7B }} \\
\midrule
MiniCPM-V-2.6~\cite{yao2024minicpm} & 48.1 & - & 75.1 & 62.8 & - & 57.5 & 90.8 & 852 & 82.1 & 65.7 \\
Qwen2-VL-7B~\cite{wang2024qwen2} & 50.1 & 74.4 & 78.0 & 67.0 & 56.5 & 60.7 & 94.5 & 856 & 83.0 & 69.0 \\
InternVL2-5-7B~\cite{chen2024expanding} & 52.9 & 79.6 & 83.0 & 70.3 & 57.4 & 63.9 & 95.7 & 864 & 83.9 & 70.4 \\
InternVL2-8B~\cite{chen2024far} & 45.2 & 71.6 & 78.1 & 66.1 & 59.1 & 61.6 & 91.6 & 794 & 83.0 & 69.7 \\
InternVL2.5-8B~\cite{chen2024expanding} & 50.1 & 78.4 & 83.1 & 71.0 & 59.1 & 62.8 & 93.0 & 822 & 84.5 & 69.7 \\
\midrule
\multicolumn{11}{c}{\textbf{2B  $<$ Model parameters $\leq$ 4B   }} \\
\midrule
Qwen2.5-VL-3B~\cite{bai2025qwen2} & 46.3 & 73.6 & 77.4 & 65.4 & 53.1 & 55.9 & 93.9 & 797 & 81.6 & 67.6 \\
BlueLM-V-3B~\cite{lu2024bluelm} & 48.1 & - & 78.1 & 66.7 & - & 62.3 & 87.8 & 829 & 85.3 & - \\
Phi3.5-Vision-4B~\cite{abdin2024phi} & 40.5 & 68.5 & 72.1 & 59.7 & 35.2 & 47.5 & 69.3 & 599 & 77.8 & 62.2 \\
InternVL2-4B~\cite{chen2024far} & 41.9 & 71.1 & 75.8 & 60.7 & 52.1 & 54.3 & 89.2 & 788 & 78.9 & 63.9 \\
InternVL2.5-4B~\cite{chen2024expanding} & 46.3 & 75.5 & 79.3 & 64.4 & 55.3 & 54.3 & 91.6 & 828 & 81.4 & 66.9 \\
\midrule
\multicolumn{11}{c}{\textbf{Model parameters $\leq$ 2B }} \\
\midrule
LLaVA-OV-0.5B~\cite{li2024llava} & 27.9 & - & 59.6 & 55.6 & - & 37.7 & 70.0 & 565 & 57.1 & - \\
InternVL2-1B~\cite{chen2024far} & 34.0 & - & 59.7 & 61.6 & 40.2 & 45.7 & 81.7 & 754 & 64.1 & 54.3 \\
InternVL2.5-1B~\cite{chen2024expanding} & 39.0 & 60.9 & 68.4 & 57.5 & 44.2 & 50.1 & 84.8 & 785 & 69.3 & 59.0 \\
SmolVLM2~\cite{marafioti2025smolvlm} & 40.6 & - & 61.1 & 57.5 & - & 46.0 & 80.0 & 725 & 69.7 & 60.5 \\
Qwen2-VL-2B~\cite{wang2024qwen2} & 41.7 & - & 72.2 & \underline{62.6} & - & 48.0 & \textbf{90.1} & \underline{809} & 74.7 & \underline{62.4} \\
Aquila-VL-2B~\cite{gu2024infinity} & 43.0 & - & 75.2 & - & - & 48.0 & 85.0 & 772 & 75.0 & \textbf{63.0} \\
InternVL2-2B~\cite{chen2024far} & 37.9 & 66.3 & 70.2 & 57.3 & 47.3 & 50.1 & 86.9 & 784 & 74.1 & 60.0 \\
InternVL2.5-2B~\cite{chen2024expanding} & 42.6 & 70.0 & \underline{73.4} & 60.0 & {48.8} & \underline{53.7} & 88.7 & 804 & 74.9 & 60.9 \\
\rowcolor{gray!20} {MagicVL-2B (Qwen2.5-1.5B)} & \underline{47.7} & \textbf{70.9} & {71.8} & {61.4} & \textbf{49.8} & {52.7} & {87.7} & {775} & \underline{76.7} & {61.0} \\
\rowcolor{gray!20} \textbf{MagicVL-2B (Qwen3-1.7B)} & \textbf{50.8} & \underline{70.3} & \textbf{73.7} & \textbf{63.5} & \underline{49.1} & \textbf{57.9} & \underline{89.0} & \textbf{828} & \textbf{77.4} & {61.6} \\
\bottomrule
\end{tabular}
\caption{
Benchmark results of various VLMs. \textbf{HB}: HallusionBench, \textbf{CR}: CRPE, \textbf{MMB}: MMBench\_V11\_en, \textbf{RQA}: RealworldQA, \textbf{MME\_R}: MME\_Realworld, \textbf{MMS}: MMStar, \textbf{DocV}: DocVQA, \textbf{OB}: OCRBench, \textbf{A2D}: AI2D, \textbf{SEED}: SEED-2 Plus.}
\label{tab:benchmark_results}
\end{table*}

\subsubsection{Progressive Training Stages}

As illustrated in Figure~\ref{fig:framework}(c), we propose a \emph{four-stage curriculum learning paradigm} that incrementally strengthens the model's capability in multi-modal understanding and reasoning. Each stage is meticulously designed based on the data categorization and complexity analysis introduced in previous sections, with both training data and strategies specifically optimized for distinct learning objectives:
\textbf{Stage 1: Foundational Modality Alignment.}
We begin by aligning the visual and linguistic modalities. In this stage, the visual encoder and LLM are frozen, and only the MLP projector is updated. Training is conducted on \emph{low-complexity} image-caption pairs (10M samples), enabling the model to establish fundamental cross-modal grounding within a simplified setting.
\textbf{Stage 2: Enhanced Visual Representation.}
Subsequently, both the visual encoder and the MLP projector are jointly optimized, while the LLM remains frozen. The training set is extended to include \emph{high-complexity} image-caption pairs (23M samples), which encourages the model to learn richer visual features and more robust cross-modal representations.
\textbf{Stage 3: Generalized Multi-Modal Ability.}
At this stage, all components---the visual encoder, MLP projector, and LLM---are unfrozen for joint training. We utilize diverse multi-modal instruction-following tasks, leveraging only \emph{low-complexity} datasets (54M samples). By gradually increasing task difficulty, this stage mitigates catastrophic forgetting and cultivates generalized reasoning ability.
\textbf{Stage 4: Advanced Multi-Modal Ability.}
Finally, the model is trained on the most challenging samples (\emph{high-complexity} data spanning all tasks, 66M samples), with all components optimized jointly. This stage consolidates advanced reasoning abilities and significantly boosts performance on both general and fine-grained tasks, particularly in real-world mobile scenarios.
This progressive, complexity-aware curriculum facilitates a seamless transition from fundamental modality alignment to advanced multi-modal reasoning. As a result, the model acquires both robust generalization and strong task-specific capabilities.

\section{Experiments}\label{experiment}

In this section, we conduct a series of experiments to validate the effectiveness of our proposed approaches and to demonstrate the capabilities of MagicVL-2B in terms of benchmark accuracy and deployment efficiency. Unless otherwise specified, MagicVL-2B refers to the model using Qwen3-1.7B as its language model.

\begin{table*}[ht]
\begin{tabular}{lcccccc}
\midrule
\textbf{Model   Name} &  \textbf{Processor} & \textbf{Model Loading} & \textbf{ViT latency} & \textbf{LLM latency} & \textbf{Throughput} \\ \midrule
InternVL2.5-2B~\cite{yao2024minicpm} & Snapdragon 8 Elite & 1.04 s & 0.90 s & 2.0 s & 14.3 token/s \\ \midrule
\rowcolor{gray!20} MagicVL-2B & Snapdragon 8 Elite & \textbf{1.01}  s & \textbf{0.09}  s & \textbf{1.7 s} & \textbf{23.9} token/s \\ \midrule
\end{tabular}
\caption{Deployment efficiency comparison with InternVL2.5-2B. MagicVL-2B achieves lower ViT and LLM inference latency as well as higher throughput compared to InternVL2.5-2B.}
\label{tab:deploy_internvl}
\end{table*}

\subsection{Training Setting}

The training pipeline for MagicVL-2B is structured into four progressive stages, in accordance with our curriculum learning paradigm. Several key hyperparameters remain consistent across all stages, including a packed batch with a maximum token length of 16,384 and up to 48 images, a maximum of 24 dynamic patches, the AdamW optimizer~\cite{loshchilov2017decoupled}, and a cosine learning rate decay schedule. All training experiments are conducted on 128 NVIDIA A800 80G GPUs. Stage-specific hyperparameters are incrementally adjusted to progressively enhance the model's capability, maintaining a balance between performance and training efficiency. Specifically, \textbf{Stage 1:} a learning rate of $2 \times 10^{-4}$, 100 warmup steps, and 65k training steps; \textbf{Stage 2:} a learning rate of $1 \times 10^{-5}$, 100 warmup steps, and 90k training steps; \textbf{Stage 3:} a learning rate of $4 \times 10^{-5}$, a warmup ratio of 0.03, and 140k training steps; \textbf{Stage 4:} a learning rate of $4 \times 10^{-5}$, a warmup ratio of 0.03, and 250k training steps.

\subsection{Comparison with State-of-the-art}

We evaluate the performance of our MagicVL-2B model against a comprehensive selection of state-of-the-art multimodal models across multiple benchmarks, as summarized in Table~\ref{tab:benchmark_results}. For fair comparison, models are grouped according to their parameter scale. Within the $\leq$2B parameter regime, MagicVL-2B consistently achieves the highest scores on the majority of benchmarks, including HallusionBench (50.8), MMBench (73.7), RealworldQA (63.5), MMStar (57.9), OCRBench (828), and AI2D (77.4), demonstrating strong capabilities in both vision-language reasoning and real-world understanding tasks. Notably, MagicVL-2B surpasses several larger models, particularly on HallusionBench and OCRBench, where it even outperforms models with over 7B parameters. These results highlight the efficiency and effectiveness of MagicVL-2B, especially in light of its compact model size. The substantial performance improvements underscore the strength of our design in developing lightweight yet powerful multimodal models.

\subsection{Ablation Study}
\begin{table}[t]

\centering

\begin{tabular}{lccccccc}
\toprule
Models & Token & TextVQA   \\
\midrule
InternVL2.5-2B ~\cite{yao2024minicpm} & 0.82 M & 74.3  \\
MagicVL-2B & 0.51 M & \textbf{74.5}   \\

\bottomrule
\end{tabular}
\caption{Ablation study of dynamic resolution with different  visual tokens comsumption on TextVQA.}
\label{tab:dynamic}
\end{table}

\subsubsection{Effectiveness of Dynamic High Resolution}
We compare the dynamic high-resolution method in our MagicVL-2B with InternVL on the TextVQA dataset, which is specifically designed to evaluate a model’s OCR and multi-modal reasoning capabilities in complex scenarios. 
As shown in Table~\ref{tab:dynamic}, MagicVL-2B reduces the total number of tokens by approximately 37.8\% (0.52 M vs 0.81 M) during the evaluation, while also achieving improved accuracy (74.5\% vs 74.3\%). 
These results demonstrate that our dynamic high resolution method can significantly reduce computatioddnal costs while still preserving detailed information.


\subsubsection{Effectiveness of Curriculum Learning Pre-Training}
We conduct an ablation study on MagicVL-2B to assess the impact of curriculum learning, as presented in Table~\ref{tab:CL_ablation}. The first baseline trains on all available data in the initial stages (excluding Data C. and Prog T.), resulting in the lowest accuracy among the compared methods. The second baseline introduces Data C., utilizing caption data in stages 1–2 and mixed data in stages 3–4, which improves the model’s fundamental multimodal capabilities, particularly for general vision-language understanding and hallucination tasks such as CR and HB. Our full curriculum learning strategy further enhances performance across all datasets, with notable improvements on challenging and fine-grained tasks, including multi-image, OCR, and reasoning benchmarks such as MME\_R, DocV, and MMS. These findings demonstrate that curriculum learning substantially boosts cross-modal understanding and generalization in lightweight models, effectively narrowing the performance gap with larger models and enabling more efficient training and deployment.


\begin{table}[t]
\centering

\begin{tabular}{cccccc}
\toprule
Data C. & Prog T. & HB & CR & MME\_R & DocV \\
\midrule
\XSolidBrush   & \XSolidBrush  & 49.5 & 69.2 & 48.2 & 87.5    \\
\CheckmarkBold   & \XSolidBrush & 50.3 & 70.0 & 48.4 & 88.1    \\
\CheckmarkBold    & \CheckmarkBold   & \textbf{50.8} & \textbf{70.3} & \textbf{49.1} & \textbf{89.0}   \\

\bottomrule
\end{tabular}
\caption{Ablation study of curriculum learning training. Data C.: data categorization, Prog T.: progressive training. \textbf{HB}: HallusionBench, \textbf{CR}: CRPE, \textbf{MME\_R}: MME Realworld, \textbf{DocV}: DocVQA}
\label{tab:CL_ablation}
\end{table}
\subsubsection{Deployment Efficiency Evaluation}
We conduct a head-to-head comparison between MagicVL-2B and InternVL2.5-2B~\cite{chen2024expanding} on the same Snapdragon 8 Elite processor. As summarized in Table~\ref{tab:deploy_internvl}, MagicVL-2B demonstrates substantial improvements in deployment efficiency across various metrics. Specifically, MagicVL-2B achieves a significantly lower ViT inference latency of 0.09s, compared to 0.90s for InternVL2.5-2B, reflecting a remarkable reduction in visual feature extraction time. Furthermore, MagicVL-2B attains a throughput of 23.9 tokens/s, approximately 1.67$\times$ higher than that of InternVL2.5-2B (14.3 tokens/s), indicating a more efficient token generation process and enhanced suitability for real-time applications. These results underscore the advantages of MagicVL-2B for deployment on resource-constrained edge devices, establishing it as a compelling solution for mobile and embedded AI scenarios.


\section{Conclusion}\label{conclusion}

In summary, MagicVL-2B demonstrates that it is feasible to achieve both state-of-the-art performance and outstanding efficiency within a lightweight multimodal framework. 
By integrating an efficient visual encoder with a curriculum learning strategy, MagicVL-2B establishes a new benchmark for small-scale MLLMs, achieving strong results on challenging benchmarks while maintaining low power consumption and latency. 
These advantages underscore its practical utility for real-world deployment, particularly in resource-constrained environments. 
We believe that MagicVL-2B paves the way for further research into scalable and efficient multimodal models, and serves as a robust foundation for deployment across diverse devices and application scenarios.

\bibliography{aaai2026}

\end{document}